\documentclass[10pt,twocolumn,letterpaper]{article}

\usepackage{cvpr}
\usepackage{times}
\usepackage{epsfig}
\usepackage{graphicx}
\usepackage{amsmath}
\usepackage{amssymb}

\usepackage{wrapfig}
\usepackage{tabulary}

\usepackage{microtype}
\usepackage{subfigure}
\usepackage{booktabs} 

\usepackage[normalem]{ulem} 

\usepackage{kotex} 
\usepackage{adjustbox}
\usepackage{tabularx}
\usepackage{multirow}
\makeatletter
\newcommand{\thickhline}{%
	\noalign {\ifnum 0=`}\fi \hrule height 1.3pt
	\futurelet \reserved@a \@xhline
}

\usepackage[pagebackref=true,breaklinks=true,letterpaper=true,colorlinks,bookmarks=false]{hyperref}

\cvprfinalcopy 

\ifcvprfinal\fi
\begin{document}

\title{Fine-Grained Neural Architecture Search}

\author{Heewon Kim$^\dagger$ \hspace{0.5cm} Seokil Hong$^\dagger$ \hspace{0.5cm} Bohyung Han$^\dagger$ \hspace{0.5cm} Heesoo Myeong$^\ddagger$ \hspace{0.5cm} Kyoung Mu Lee$^\dagger$\\
$^\dagger$Computer Vision Lab \& ASRI, Seoul National University  \hspace{0.5cm} $^\ddagger$ Qualcomm Korea YH   \\
{\tt\small \null \hspace{1.1cm} \{ghimhw, hongceo96, bhhan, kyoungmu\}@snu.ac.kr \hspace{0.7cm} hmyeong@qti.qualcomm.com  \hspace{\fill}  \null }
}

\maketitle

\begin{abstract}

We present an elegant framework of fine-grained neural architecture search (FGNAS), which allows to employ multiple heterogeneous operations within a single layer and can even generate compositional feature maps using several different base operations.
FGNAS runs efficiently in spite of  significantly large search space compared to other methods because it trains networks end-to-end by a stochastic gradient descent method.
Moreover, the proposed framework allows to optimize the network under predefined resource constraints in terms of number of parameters, FLOPs and latency.
FGNAS has been applied to two crucial applications in resource demanding computer vision tasks---large-scale image classification and image super-resolution---and demonstrates the state-of-the-art performance through flexible operation search and channel pruning.

\end{abstract}


\section{Introduction}
\label{sec:introduction}

Deep convolutional neural networks (CNNs) have recently achieved great success in various fields including computer vision, natural language processing, pattern recognition, bioinformatics, and many others.
However, the arbitrary complexity of target problems and the requirement of extensive hyperparameter search make it inevitable to manually explore the ideal deep network architectures customized for the given tasks.
Consequently, neural architecture search (NAS) approaches have been studied actively, and the models identified by the NAS techniques~\cite{zoph2016nas,pham2018enas,tan2018mnas, he2018amc} started to surpass the performance of the traditional deep neural networks~\cite{simonyan2014vgg,he2015resnet,huang2016dense} designed by human.
Despite such successful results, it is still a challenging problem to optimize deep neural networks even by sophisticated AutoML techniques because the search space of the existing NAS methods is limited while their search cost is high.

\begin{figure}[t]
	\centering
	\includegraphics[width=0.92\linewidth]{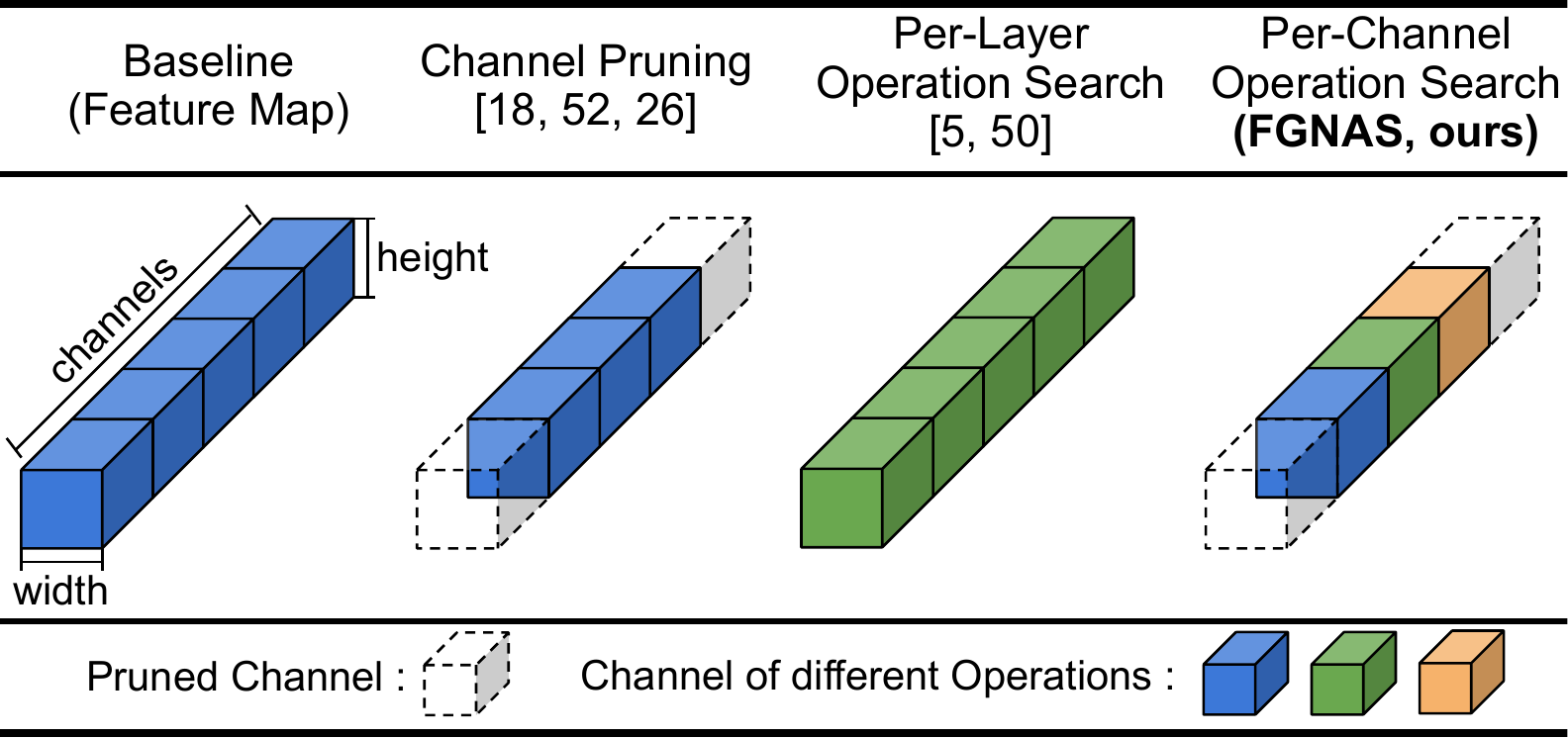} \\
	{\small (a) Flexibility Comparison of Searched Architectures }\\ \vspace{0.3cm}
	\includegraphics[width=0.92\linewidth]{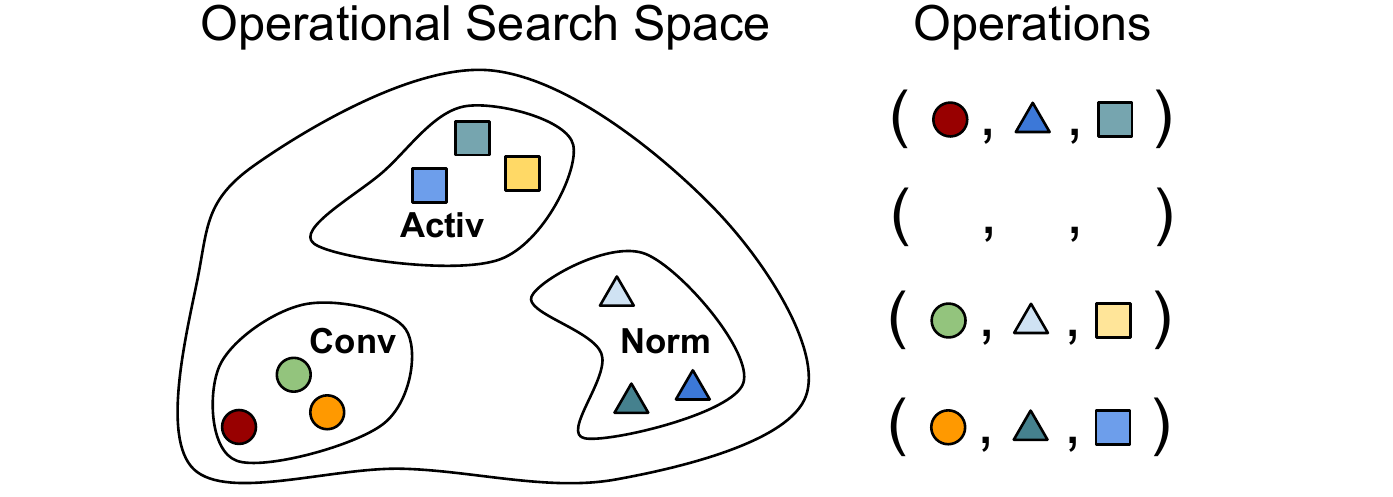} \\
	{\small (b) Our Definition of Convolution Operation} \\ \vspace{0.2cm}
	\caption{ 
		{
		Overview of our search method. 
		(a) FGNAS performs operation search for each of channels, which leads to more flexible network architectures than other approaches. The example illustrates the architecture search results in a single layer with 5 channels, where color encodes an operation. Note that the flexibility of the output models depends on search granularity of individual algorithms. 
		(b) In this paper, we define an {\it operation} as a sequence of a convolution, a normalization, and an activation function, denoted by $(\cdot,\cdot,\cdot)$. Note that the operation search even includes channel pruning, which is equivalent to no-operation.}
	}
	\vspace{-0.4cm}
	\label{fig_overview}
\end{figure}

\begin{table*}[t]
	\centering
	\caption{{Comparisons of automated neural architecture search (or pruning) techniques. Our algorithm conceptually has the largest search space among all the compared methods, and efficiently optimizes candidate models by gradient-based search strategies.   }}
	\vspace{0.2cm}\hspace{-0.21cm}
	\scalebox{0.725}{
		\begin{tabular}{l  c  c c c c c c} 
			\toprule
			& & AMC~\cite{he2018amc} & NetAdapt~\cite{yang2018netadapt} &  Huang et al.~\cite{Huang2018learning}&  MnasNet~\cite{tan2018mnas} & ProxylessNAS \& FBNet~\cite{cai2018proxylessnas,Wu2019fbnet}  & \textbf{FGNAS (Ours)} \\ 
			\midrule
			\multirow{1}{*}{Structure search} & Prune channels &  \checkmark  & \checkmark & \checkmark&   & & \checkmark \\ \\[-1em]
			\hline \\[-0.8em]
			\multirow{3}{*}{Operation search} & Find efficient operations & &  & &  \checkmark &\checkmark& \checkmark \\ \\[-1em]
			& Layer-wise optimization & \checkmark &\checkmark& \checkmark &  & \checkmark& \checkmark \\
			& Channel-wise optimization & \checkmark &\checkmark&  \checkmark&  & & \checkmark \\
			\hline \\[-0.8em]
			\multicolumn{2}{c}{Optimization method}& RL & trial-and-error & policy-gradient & RL & gradient-based & gradient-based\\
			\bottomrule
		\end{tabular}}
		\label{table:acceleration_approaches}
	\end{table*}

Researchers have aimed to develop flexible and scalable NAS techniques with large search spaces and identify the unique models different from the manually designed structures~\cite{zoph2016nas}.
However, NAS methods often suffer from huge computational cost and reduce their search space significantly for practical reasons.
For example, \cite{zoph2016nas, baker2017designing, pham2018enas,liu2018pnas,liu2018darts} search for two cells as basic building blocks to construct full models by stacking them. 
To tackle the redundancy between the cells and increase the diversity of full models, MnasNet~\cite{tan2018mnas} adopts smaller search units, blocks, than cells.
Recently, FBNet~\cite{Wu2019fbnet} and ProxylessNAS~\cite{cai2018proxylessnas} reduce their search units further to individual layers.
Although the resulting models are more flexible by decreasing the granularity of search units and increasing the diversity of the generated models through their composition, those methods are limited to allocating a single operation per layer and the operation configurations of the whole network are proportional to the number of layers.

On the contrary, we present a flexible and scalable neural architecture search algorithm.
The search unit of our algorithm is channel, which is even smaller than layer; each channel chooses a different operation\footnote{We define a series of a convolution, normalization and activation function application by an operation.}, which also includes {\it no-operation}, equivalent to channel pruning.
This kind of search strategy improves the flexibility of resulting models because it is possible to generate a large number of configurations even within a single layer, which increase exponentially by adding layers.
Such an extremely flexible framework incurs small overhead, which allows to maintain various operations for search and increase search space significantly.
Figure~\ref{fig_overview} illustrates the proposed fine-grained neural architecture search (FGNAS) approach, where our per-channel search algorithm generates a feature map given by a composition of multiple operations and also reduces the number of channels by pruning.

FGNAS is trained to maximize the validation accuracy efficiently and stably by a stochastic gradient descent method.
Moreover, it is convenient to regularize individual channels by incorporating FLOPs and latency into the training objective.
Therefore, the proposed algorithm has a great deal of flexibility and scalability to maximize the accuracy of searched models while facilitating to consider various aspects for optimization.
Our overall contribution is summarized as follows:
	\begin{itemize}
		\item We propose a flexible and scalable fine-grained neural architecture search algorithm, which allows to perform per-channel operation search including channel pruning efficiently and optimize end-to-end by a stochastic gradient descent method.
		\item Our framework deals with diverse objectives of neural architecture search such as number of parameters, FLOPs and latency, in addition to accuracy, conveniently.
		\item	The resulting models from our algorithm achieve outstanding performance improvements with respect to various evaluation metrics in image classification and single image super-resolution problems.
	\end{itemize}

	The rest of this paper is organized as follows.
	We first discuss existing works related to deep neural network optimization and neural architecture search in Section~\ref{sec:related}.
	Section~\ref{sec:proposed} describes the proposed algorithm in details including training methods and Section~\ref{sec:experiment} presents experimental results in comparison to the existing methods.


\section{Related Work}
\label{sec:related}
This section describes existing efficient convolution network designs and neural architecture search techniques in details. 
Table~\ref{table:acceleration_approaches} presents the snapshot of the algorithms discussed in this section.

\vspace{-0.2cm}
\paragraph{Efficient Convolution Networks}
Designing compact convolutional neural networks has been an active research problem in the last few years.
While the hand-crafted models achieve efficient convolutional operations by revising network structures~\cite{landola2016squeezenet,howard2017mobilenet,zhang2018shufflenet,sandler2018mobilenetv2,huang2017condensenet}, the simple rule-based network quantization~\cite{han2015deep_compression} and pruning techniques~\cite{han15learning,Denton14exploiting,dong2017more,han2015deep_compression, molchanov2016pruning, liu2017slimming,li2016pruning,He_2017_ICCV} reduce the redundancy of deep and complex pretrained models successfully.
Recent pruning methods automatically remove filters and/or activations using reinforcement learning~\cite{he2018amc}, trial-and-error~\cite{yang2018netadapt}, and policy-gradient~\cite{Huang2018learning}. 
They optimize a network in a layer-by-layer fashion, which is inefficient in dealing with inter-layer relationships, while our FGNAS optimizes all layers jointly using a gradient-based method.

\begin{figure*}[t]
	\centering
	\includegraphics[width=0.85\linewidth]{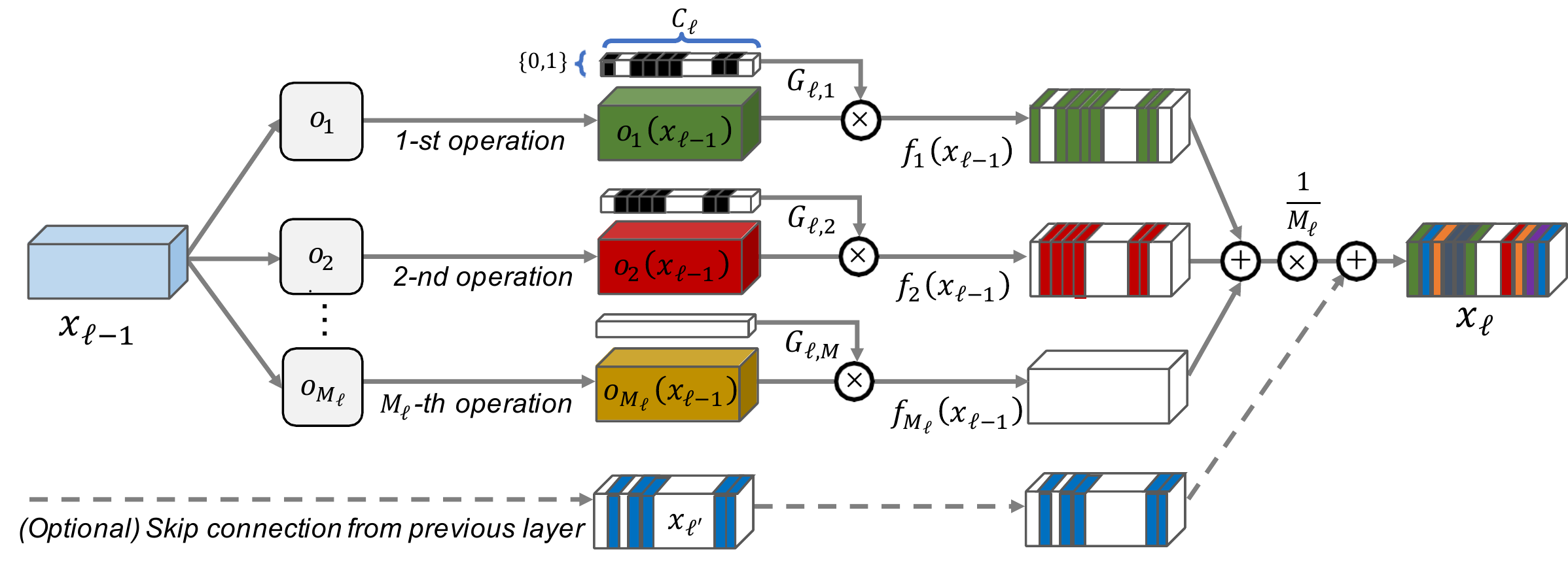}
	\caption{
	The proposed efficient per-channel operation search framework using binary mask $G_{\cdot,\cdot}$. 
	The $i$-th operation $o_i(\cdot)$ produces a tensor with $C_\ell$ channels from the input feature map $x_{\ell-1}$ in the $\ell$-th layer. The binary vector $G_{\ell,i}$ selects a subset of channels in the output tensor $o_i(x_{\ell-1})$, and each masked tensor $f_i(x_{\ell-1})$ and the optional skip-connection feature map $x_{\ell'}$ of the $\ell'$-th layer are aggregated to the final output feature map $x_{\ell}$ of the $\ell$-th layer.
	This framework facilitates efficient search for flexible architectures. 
	Note that we just need to learn the binary masks for neural architecture search, which are implemented by gating functions.
	Figure~\ref{fig_channel} illustrates the details about forward and backward passes along the gating functions.}
	\label{fig_framework}
	\vspace{-0.2cm}
\end{figure*}

\vspace{-0.2cm}
\paragraph{Neural Architecture Search (NAS)}
Automatic architecture search techniques conceptually have more flexibility in the identified models than the hand-crafted methods.
NASNet~\cite{zoph2016nas} and MetaQNN~\cite{baker2017designing} adopt reinforcement learning for non-differential optimization.
ENAS~\cite{pham2018enas} employs a RNN controller to search for the optimal model by drawing a series of sample models and maximizing their expected reward, while PNAS~\cite{liu2018pnas} performs a progressive architecture search by predicting accuracy of candidate models.
Evolutionary search~\cite{real2018evolution} employs a tournament selection; although it is the first algorithm to surpass the state-of-the-art classification accuracy, it requires significantly more computational resources.
DARTS~\cite{liu2018darts} relaxes the discrete architecture representation to a continuous one and addresses scalability issue by making the objective function differentiable.
MnasNet~\cite{tan2018mnas} and DPP-Net~\cite{dong2018dppnet} are optimized with respect to the accuracy and run-time via reinforcement learning and performance predictor.
EfficientNet~\cite{tan2019efficientnet} improves network efficiency by simply scaling depth, width, and resolution of backbone network.
MobileNetV3~\cite{howard2019mobilenetv3} adopts block-wise search~\cite{tan2018mnas} with layer-wise pruning~\cite{yang2018netadapt} and presents a novel architecture design with Squeeze-and-Excitation~\cite{Hu18squeeze}.
Recently, multiple choice gating function is often adopted for differentiable and multi-objective search techniques.
ProxylessNAS~\cite{cai2018proxylessnas} and FBNet~\cite{Wu2019fbnet} search for efficient convolution operations in each layer.
MixConv~\cite{mingxing2019MixConv} finds a new depth-wise convolution operation that has multiple kernel sizes within a layer.
Our FGNAS presents per-channel convolution operation search, which constructs maximally flexible layer configurations as illustrated in Figure~\ref{fig_overview} and runs efficiently through a differentiable optimization.


\section{Proposed Algorithm}
\label{sec:proposed}

This section first presents our efficient search formulation via binary masking, and discusses our gating function that allows to perform the end-to-end differentiable search.
Then, we present the objective function of our algorithm based on resource regularizer, which directly penalizes each channel, and describes the exact search space.

\subsection{Formulation of Operation Search}
\label{method:formulation}

\begin{figure}[t]
	\centering
	\includegraphics[width=0.75\linewidth]{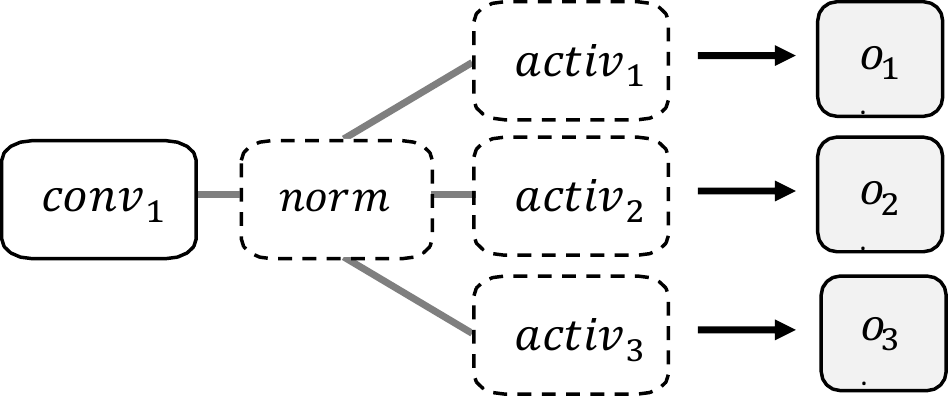}
	\caption{The structure of multiple operations from a convolution.
	The multiple operations share the feature map of normalization to reduce search cost.  Since an operation consists of a sequence of a convolution, a normalization, and an activation function, dashed boxes may be omitted depending on the backbone networks.
	}
	\vspace{-0.2cm}
	\label{fig_operation_definition}
\end{figure}

\begin{figure*}[t]
	\centering
		\includegraphics[height=0.25\linewidth]{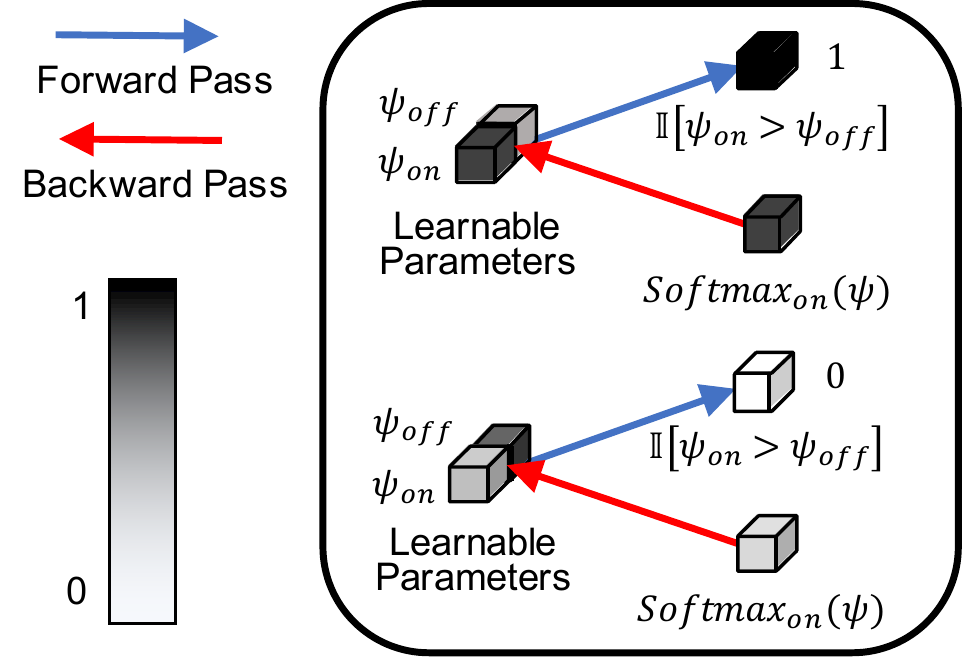} \hfill
		\includegraphics[height=0.25\linewidth]{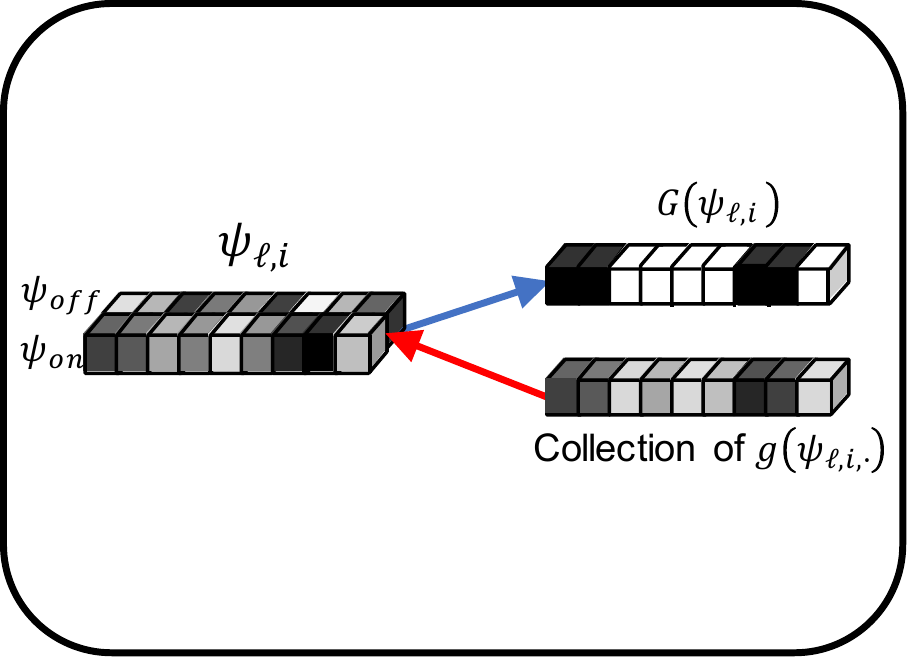} \hfill
		\includegraphics[height=0.25\linewidth]{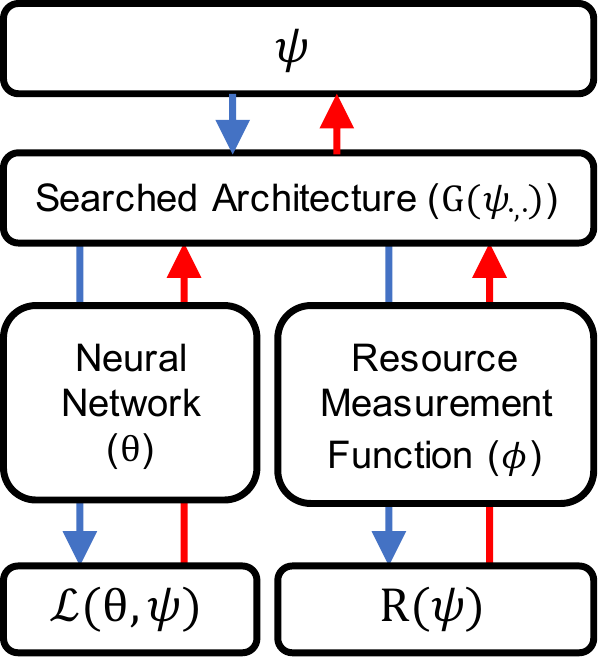} \\ \vspace{-0.3cm}
		{\small \null \hspace{0.1\linewidth} (a) Per-Channel Gating Function $g(\psi)$\hspace{0.06\linewidth} (b) Relaxed Binary Mask $G(\psi_{\ell,i})$ \hspace{0.08\linewidth} (c) Resource Regularizer $R(\psi)$ \hspace{\fill}  \null }
	\vspace{-0.2cm}
	\caption{
	Illustration of forward and backward passes to optimize the gating function parameters $\psi$.
	(a) Gating function $g(\cdot)$ produces a binary value in the forward pass and a softmax probability in the backward pass for gradient-decent optimization.
	(b) The collection of gating functions $G(\psi_{\ell,i})$ is a relaxed version of $G_{\ell,i}$ in~\eqref{eq:op_prune}.
	(c) $\psi$ controls searched architectures by determining active channels in the forward pass.
	During the gradient-decent optimization procedure in the backward pass, resource regularizer $R(\psi)$ plays a role to penalize a channel with high resource consumption, while the task-specific loss $\mathcal{L}(\theta,\psi)$ attempts to keep the channel alive if it performs well in the target task.}
	\label{fig_channel}
	\vspace{-0.2cm}
\end{figure*}

Although FGNAS has a large search space and generates flexible output models, a critical concern is how to perform NAS efficiently through proper configuration of the search space.
To tackle this challenge, FGNAS constructs a feature map using a composition of multiple operations as illustrated in Figure~\ref{fig_framework}, where the composition allows to generate a large number of virtual operations and increase the flexibility of searched models.
Given an input tensor in the $\ell$-th layer, denoted by $x_{\ell-1}$, the output of the layer, $x_{\ell}$, is expressed as
\begin{equation}\label{eq:op_search}
x_{\ell} = {1\over M_\ell} \sum_{i=1}^M f_i(x_{\ell-1}), 
\end{equation}
where $M_\ell$ is the number of operations at $\ell$-th layer considered in our search and
\begin{equation}\label{eq:op_prune}
 f_i(x_\ell)  \equiv  G_{\ell, i} \odot o_i(x_{\ell-1}).
\end{equation}
Note that $G_{\ell, i} \in \mathbb{R}^{C_\ell}$ is a binary vector, $o_i(\cdot)$ represents the $i$-th operation producing a tensor with $C_\ell$ channels, and $\odot$ denotes the channel-wise binary masking operator.
In other words, the output tensor $x_{\ell}$ is given by the average of $M$ masked tensors,
where the mask of each tensor is learned by our search algorithm, which also allows channel pruning by masking out the same channels in all output tensors.
In addition to the operation search, we consider the identity connections from a preceding layer optionally, which derives the modification of \eqref{eq:op_search} as
\begin{equation}\label{eq:op_search2}
x_{\ell} = x_{\ell'} + {1\over M_\ell} \sum_{i=1}^{M_\ell} G_{\ell, i} \odot o_i(x_{\ell-1}),
\end{equation}
where $x_{\ell'}~(\ell' < \ell)$ denotes the feature map from which the identify connection is originated.

In our algorithm, each operation $o_i(\cdot)$ is defined by a series of convolution, normalization, and activation function application as illustrated in Figure~\ref{fig_overview}~(b). 
Figure~\ref{fig_operation_definition} presents our efficient operation structure to increase the number of operations with little additional cost because all the three operations in Figure~\ref{fig_operation_definition} share the previous feature map of a normalization.
For some parts of backbone networks that convolutional layers are not followed by normalization and activation function layers, an operation is actually equivalent to a convolution.

\subsection{Per-Channel Differentiable Gating Functions}
\label{method:gate_search}
To relax the binary mask $G_{\cdot,\cdot}$ in~\eqref{eq:op_prune},
we introduce a relaxed gating function $g(\cdot)$, and define a collection of the gating functions, denoted by $G(\cdot)$, as
\begin{equation}\label{eq:large_g}
G\left(\psi_{\ell,i}\right) = \left[g\left(\psi_{\ell,i,1}\right), \dots,g\left(\psi_{\ell,i,C_\ell}\right)\right]^\top,
\end{equation}
where $\ell$ and $i$ denote the layer and operation index, respectively, and $C_\ell$ is the number of channels.
A relaxed gating function for each channel parametrized by $\psi = [\psi_\text{on}, \psi_\text{off}]^\top \in \mathbb{R}^2$ is given by
\begin{equation}\label{eq:gating_function}
g(\psi) = 
\begin{cases}
\mathbb{I}\left[\psi_\text{on}>\psi_\text{off}\right]& \mbox {if forward} \\
\text{softmax}_\text{on}(\psi) & \mbox{if backward},
\end{cases}
\end{equation}
where $\mathbb{I}\left[\cdot\right]$ is an indicator function that returns 1 when its input is true and 0 otherwise, and $\text{softmax}_\text{on}$ denotes the value corresponding to $\psi_\text{on}$ dimension after applying a softmax function.
Figure~\ref{fig_channel}~(a) and (b) illustrate $g(\psi)$ and $G(\psi_{\ell,i})$, respectively.

Using the relaxed gating function, we reformulate the channel-wise tensor masking in \eqref{eq:op_prune} as
\begin{equation}\label{eq:op_prune_relax}
 f_i(x_{\ell-1}; \psi)  \equiv G\left(\psi_{\ell,i}\right)  \odot o_i(x_{\ell-1}).
\end{equation}
This relaxed gating function allows to update the architecture by a gradient-decent optimization method because the backward function is differentiable.

\subsection{Resource Regularizer on Channels}
\label{method:run-time}

The proposed approach aims to maximize the accuracy of a target task and minimize the resource usage of the identified model.
Hence, our objective function is composed of two terms; one is the task-specific loss and the other is a regularizer penalizing overhead of networks such as parameters, FLOPs, and latency.
To search for operations per channel, the proposed regularizer computes the amount of resource usage of a channel, which changes over iterations due to the gradual update of architectures.
Figure~\ref{fig_channel} (c) illustrates overview of the resource regularizer, and the rest of this section discusses the details.

Let $\mathcal{L}(\cdot, \cdot)$ denote a loss function for an arbitrary task\footnote{In our work, the tasks are image classification and super-resolution.} and $\mathcal{R}(\cdot)$ be a differentiable regularizer that estimates the resources of the current model identified by our search algorithm.
Then, the objective function is formally given by
\begin{equation}\label{formulation8}
\min_{\theta,\psi} \mathcal{L}(\theta, \psi) + \lambda \cdot \mathcal{R}(\psi),
\end{equation}
where $\theta$ and $\psi$ are learnable parameters in the neural networks and the gating functions $g(\cdot)$, respectively, and $\lambda$ is the hyper-parameter balancing the two terms.
Specifically, the regularizer $\mathcal{R}\left(\psi\right)$ is given by
\begin{align}\label{formulation9}
&\mathcal{R}\left(\psi \right) 
= \sum_{\ell=1}^L \sum_{i=1}^{M_\ell} 
\phi_{\ell,i}^*\left(\gamma\left(\psi_{\ell-1}\right),\gamma\left(\psi_{\ell,i}\right)\right), 
\end{align}
where $\phi_{\ell, i}^*(\cdot,\cdot)$ is a resource measurement function of the $i$-th operation, $* \in \{\text{parameters},\text{FLOPs}, \text{latency} \}$ indicates the type of the resources, and $L$ is the number of layers.
Note that $\gamma(\psi_{\ell-1})$ and $\gamma(\psi_{\ell,i})$ are the number of input/output channels of the $i$-th operation, respectively, and they are differentiable via gating function, defined as
\begin{equation}\label{eq:num_op_ch}
\gamma\left(\psi_{\ell,i}\right) \equiv
\lVert G\left(\psi_{\ell,i}\right) \rVert_1 
\end{equation}
and
\begin{equation}\label{eq:num_layer_ch}
\gamma\left(\psi_{\ell-1}\right) \equiv
\left|\left|  h\left(\sum_{i=1}^M G\left(\psi_{\ell-1,i}\right) \right) \right|\right|_1,
\end{equation}
where $\lVert \cdot \rVert_1$ denotes $\ell_1$ norm of a vector.
The function $h(\cdot)$ produces a binary vector, valued 1 for non-zero elements of the input vector in the forward pass, but is an identity function in the backward pass. 
The skip connection from an earlier layer affects \eqref{eq:num_layer_ch} because we need to consider an extra term for the summation.

On the other hand, $\phi_{\ell, i}^\text{parameters}$ and $\phi_{\ell, i}^\text{FLOPs}$ are well-defined functions of convolution kernel sizes, number of channels, feature map resolution, etc.
They are differentiable with respect to the number of active channels by the definitions in~\eqref{eq:num_op_ch} and \eqref{eq:num_layer_ch}.
However, it is not straightforward how to define the latency measurement function $\phi_{\ell, i}^\text{latency}$ on specific devices such as Google Pixel 1 and Samsung Galaxy S8.
We address this problem by fitting affine functions of the relation between latency and FLOPs, which are parameterized by $\phi$; it turns out that the convolution operations present strong correlations between latency and FLOPs in a particular condition provided by the combination of input feature map size, kernel size, stride, convolutional groups and so on.
By approximating latency as a function of FLOPs, \eqref{formulation9} with $\phi_{\ell, i}^\text{latency}$ naturally penalizes all channels to minimize the run-time of networks.

\subsection{Search Space}
\label{method:search_space}
FGNAS searches for an operation in each channel; the granularity of architecture search is as small as a channel.
Consequently, the possible combinations of operations in FGNAS is significantly more than those of any other NAS techniques.
Specifically, the search space in a single layer is $2^{M_\ell C_\ell}$, where $M_\ell$ is the number of operations and $C_\ell$ is the number of channels at $\ell$-th layer, while it has minor variations depending on the network configurations ({\it e.g.}, existence of skip connections).
This is truly beyond the comparable range to other approaches because most of the NAS techniques are limited to adopting a per-layer search strategy and exploring few building blocks instead of directly optimizing the whole model.

Table~\ref{table:search_space} illustrates the search space of operations in our search algorithm.
The backbone networks for image classification include VGG, ResNet, DenseNet, EfficientNet, and MobileNetV2, while EDSR is employed for image super-resolution.
Note that we insert a 1$\times$1 convolution operation after an identity connection to reduce the number of input channels to the first convolution operation of a residual (or dense) block.

\begin{table}[t]
	\centering
	\caption{The search space of operations.
	}
	\label{table:search_space}
	\vspace{0.2cm}\hspace{-0.3cm}
	\scalebox{0.85}{
	\setlength\tabcolsep{3pt}
	\begin{tabular}{l | c } 
        	\toprule[1.5pt]
          Factor & Search Space \\[0.5ex] 
	\midrule[1.0pt]
         Convolution types & Normal, Depth-wise \\
         Convolution kernel sizes & 1, 3, 5, 7, 9, 11 \\
    	Normalization method & BN \\
     	Activation functions & ReLU, PReLU, tanh \\
    	The number of channels &0, 1, 2, $\sim$, $C_{\ell-1}$, $C_\ell$ \\
        \bottomrule[1.5pt]
	\end{tabular}} 
	\\ \vspace{-0.2cm}
\end{table}


\begin{table*}[t]
\centering
\caption{Comparison with state-of-the-art architectures on CIFAR-10.}
\vspace{0.2cm}\hspace{-0.15cm}
\scalebox{0.85}{
\begin{tabular}{l  l rrr}
	\toprule[1.5pt]
     Model & Type & Search Cost (GPU-days) & Top-1 Acc.  & Parameters  \\[0.5ex] 
     \midrule[1.0pt]
	DenseNet-BC~\cite{huang2016dense} 					& manual			& -		& 96.5 \%	&  25.6 M \\
	Hierarchical Evolution~\cite{liu2018hierarchical} 			& evolution		& 300 	& 96.3 \%	& 15.7 M	\\
	P-DARTS (large)~\cite{Chen2019ICCV} + cutout			& gradient-based	& 0.3 	& 97.8 \%	& 10.5 M	\\
	ProxylessNAS-G~\cite{cai2018proxylessnas} + cutout			& gradient-based	& 4.0 	& 97.9 \%	& 5.7	 M\\
	ENAS~\cite{pham2018enas} + cutout 					& RL				& 0.5		& 97.1 \%	& 4.6	 M\\
	EfficientNet-B0~\cite{tan2019efficientnet}					& model scaling	& -	 	& 98.1 \%	& 4.0	 M\\
	\hline
	\textbf{EfficientNet-B0-FGNAS (Large) + cutout}				& gradient-based 	& \textbf{0.1}		& \textbf{98.2 \%}		& \textbf{3.9 M}	\\
	\midrule[1.0pt]
	P-DARTS~\cite{Chen2019ICCV} + cutout				& gradient-based	&  \textbf{0.3} 	& 97.5 \%	& 3.4	 M	\\
	NASNet-A~\cite{Zoph2017LearningTA} + cutout 				& RL	 			& 1800 	& 97.4 \% 	& 3.3	 M \\
	DARTS~\cite{liu2018darts}  (first order) + cutout 		& gradient-based	& 1.5		& 97.0 \%	& 3.3	 M\\
	DARTS~\cite{liu2018darts} (second order) + cutout	&gradient-based 	& 4		& 97.2 \%	& 3.3	 M\\
	AmoebaNet-A~\cite{real2018evolution} + cutout 			& evolution		& 3150	& 96.6 \%	& 3.2	 M\\
	PNAS~\cite{liu2018pnas}						& SMBO			& 225	& 96.6 \%	& 3.2	 M \\
	SNAS~\cite{xie2018snas} + mild constraint + cutout		& gradient-based	& 1.5		& 97.0 \%	& 2.9 M	\\
	SNAS~\cite{xie2018snas} + moderate constraint + cutout	& gradient-based	& 1.5		& 97.2 \%	& 2.8	 M \\
	AmoebaNet-B~\cite{real2018evolution} + cutout 				& evolution		& 3150	& 97.5 \%	& 2.8	 M \\
	\hline
	\textbf{EfficientNet-B0-FGNAS (Small) + cutout }				& gradient-based 	& 0.5		& \textbf{97.8 \%}		& \textbf{2.7 M}\\

	\toprule[1.5pt]
\end{tabular}}
		\label{table:cifar_search}
\end{table*}

\begin{table*}[t]
\centering
\caption{Comparison with channel pruning methods on ImageNet. $^\dagger$ is a reported result and similar latency with Multiplier (0.75) in~\cite{yang2018netadapt}.  }
\vspace{0.2cm}\hspace{-0.15cm}
\scalebox{0.85}{
\begin{tabular}{l l ll rrrr}
	\toprule[1.5pt]
     Model & Search Space & Method        & Type &Top-1 Acc. & Parameters & FLOPs& CPU   \\[0.5ex] 
     \midrule[1.0pt]
     \multirow{5}{*}{\shortstack[c]{MobileNetV2 (224) }} & No Search & Baseline  & manual & 72.0 \%  & 3.4 M & 600 M & 75 ms\\
     \cline{2-8}
     & \multirow{3}{*}{+ Channel Pruning}  & Multiplier (0.75)~\cite{sandler2018mobilenetv2}   & manual & 69.8 \%& \textbf{2.6 M} & 418 M & 56 ms\\
    & & NetAdapt~\cite{yang2018netadapt}      & trial-and-error & 70.9 \% &  - & - & $^\dagger$64 ms \\
     & & FGNAS (P)                        & gradient-based & 70.9  \% & 3.5 M & 410 M & \textbf{53 ms}\\
     & + 5$\times$5 DConv& \textbf{FGNAS}                        & gradient-based & \textbf{71.4 \%} & 3.1 M & \textbf{378 M} & \textbf{53 ms}\\
    
	\toprule[1.5pt]
\end{tabular}}
		\label{table:mobilenetv2}
		\vspace{-0.3cm}
\end{table*}

\section{Experiment}
\label{sec:experiment}
This section first presents the benchmark datasets for image classification and super-resolution tasks, and describe the implementation details of our algorithm.
Then, we present the experimental results including performance analysis. 

\subsection{Dataset}
CIFAR-10~\cite{alexcifar10} and ILSVRC2012~\cite{ILSVRC15} are popular datasets for image classification.
The former contains 50K and 10K 32$\times$32 images for training and testing in 10 classes. 
The latter consists of 1.2M training and 50K validation images in 1,000 object categories, which are a subset of ImageNet~\cite{deng2009imagenet}. 
DIV2K~\cite{agustsson2017ntire} is a training dataset for image super-resolution, which contains 800 2K images while we evaluate super-resolution algorithms on Set5~\cite{bevilacqua2012low}, Set14~\cite{zeyde2010single}, B100~\cite{martin2001b100}, and Urban100~\cite{huang2015urban100}

\subsection{Implementation Details}
\paragraph{Search steps}
The proposed algorithm searches for architectures with 4 steps; 
(1) determine a backbone network and operations for each layer,  
(2) pre-train the network without gating functions,  
(3) search for architectures by learning gating function parameters until the resource of searched architecture reaches target resource,
(4) fine-tune the searched architectures with fixed gating function parameters.

\vspace{-0.3cm}
\paragraph{CIFAR-10} The backbone network is EfficientNet-B0~\cite{tan2019efficientnet}, of which the architecture is designed for ImageNet and transferred to CIFAR-10. 
The search space is 1, 3, and 5 kernel sizes in depth-wise convolution layers and the number of channels in all layers. 
We train the model for 160 epochs with mini-batch size 128 and initial learning rate 0.01.
The resource of interest $*$ is number of parameters of networks and the hyper-parameter $\lambda$ is set to $10^{-7}$ for resource regularizer.
We use the standard SGD optimizer with nesterov~\cite{sutskever2013nesterov} and Cutout augmentation~\cite{devries2017cutout}. 
We use weight decay and the momentum of 0.0001 and 0.9, respectively.


\begin{table}[t]
	\centering
	\caption{Ablation study of search space on CIFAR-10. 
	}
	\label{table:space_analysis}
	\vspace{0.2cm}\hspace{-0.3cm}
	\scalebox{0.86}{
	\setlength\tabcolsep{3pt}
	\begin{tabular}{l l  l  rr } 
        	\toprule[1.5pt]
          Model & Search Space& Method & Top-1 Acc. & FLOPs \\[0.5ex] 
	\midrule[1.0pt]
         \multirow{4}{*}{VGG-16} & No Search &  Baseline  & 93.7 \% & 627 M \\ \cline{2-5}

         & + Channel pruning & FGNAS (P) &  \textbf{93.6 \%} & 149 M \\ 

         & + 1$\times$1 $\sim$ 11$\times$11 Conv. &FGNAS &  \textbf{93.6 \%} & 119 M  \\ 
         & + ReLU, PReLU, Tanh  & \textbf{FGNAS} & \textbf{93.6 \%} & \textbf{110 M}  \\
	\bottomrule[1.5pt]
	\end{tabular}} 
	\\ \vspace{-0.2cm}
\end{table}

\vspace{-0.3cm}
\paragraph{ImageNet}
MobileNetV2~\cite{sandler2018mobilenetv2} is the backbone network, of which the architecture has compact designed for ImageNet classification.
The search space is 3 and 5 kernel sizes in depth-wise convolution layers and the number of channels in all layers.
We train models using mini-batch size 256 with the initial learning rates are set to 0.01.
The training epochs are 400 and the learning rates are divided by 10 at 50\% and 75\% of the total number of training epochs. 
The resource of interest $*$ is latency of networks and the hyper-parameter $\lambda$ is set to 0.0012 for resource regularizer.
We evaluate our models on Google Pixel 1 CPU using Google's Tensor-Flow Lite engine.

\vspace{-0.3cm}
\paragraph{DIV2K}
The backbone network is a small version of EDSR~\cite{EDSR}, of which each layer and the architecture have 64 channels and 16 residual blocks, respectively.
The search space is ReLU, PReLU, and tanh in activation layers and the number of channels in all layers.
The model is pre-trained for 300 epochs using Adam~\cite{kingma2015adam}, where minibatch size is 16 with learning rate $10^{-4}$, patch size 96$\times$96 pixels, $\beta_1 = 0.9$, $\beta_2 = 0.999$, $\epsilon = 10^{-8}$.
The resource of interest $*$ is FLOPs of networks and the hyper-parameter $\lambda$ is set to $10^{-9}$  for resource regularizer.
The image restoration performance measures are PSNR and SSIM on Y channel of YCbCr color space with the scaling factor 2.

\begin{figure}[t]
 	\centering
	\vspace{0.0cm}\hspace{-0.3cm}
	\includegraphics[width=0.8\linewidth]{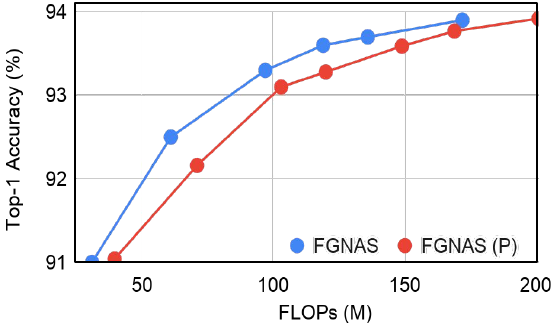}
	\caption{ Performance comparison between our algorithms from VGG-16 on CIFAR-10. FGNAS searches more efficient networks than channel pruned networks by FGNAS~(P).
	}
	\label{figure:pruning_vs_search}
\end{figure}

\begin{figure*}[t]
	\centering
	\includegraphics[width=1\linewidth]{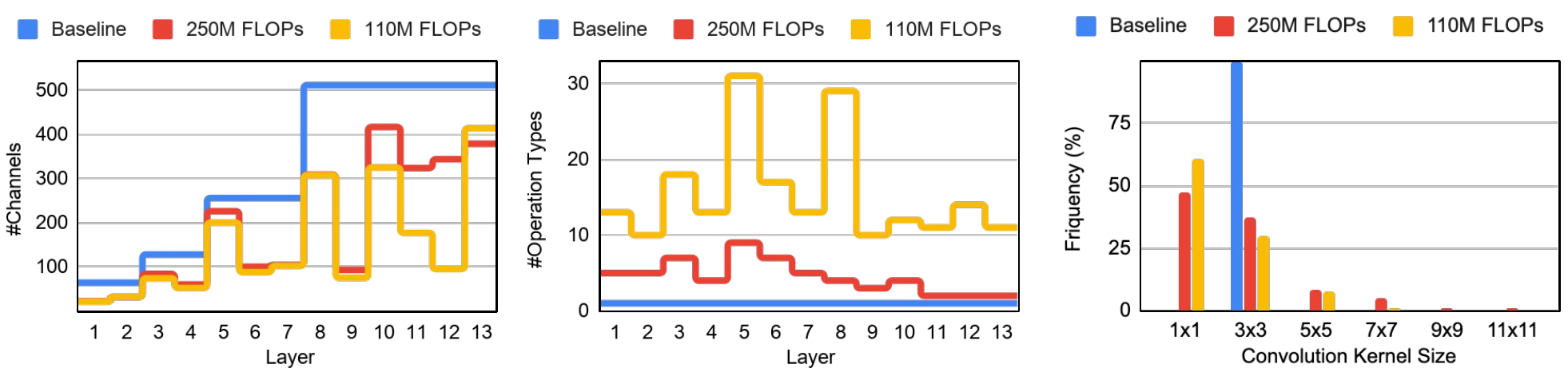} \\ \vspace{-0.5cm}
	 {\small \null \hspace{0.09\linewidth} (a) \#Channels per Layer\hspace{0.13\linewidth} (b) \#Operation Types per Layer\hspace{0.10\linewidth} (c) Frequency of Conv. Kernel Sizes \hspace{\fill}  \null}
	 \vspace{0.1cm}
	\caption{Searched architecture analysis from VGG-16 on CIFAR-10 by FGNAS. Blue, red, and yellow colors denote 627M (Baseline), 250M, 110M FLOPs networks, respectively. 
	(a) The number of channels at each layer.
	(b) The number of operation types at each layer. If more than two operations produce a channel, we account them as a new operation type for visualization.
	(c) Frequency of convolution kernel sizes in operations.}  
	\label{fig_fgnas_vgg16}
	\vspace{-0.2cm}
\end{figure*}

\begin{figure*}[t]
	\centering
	\includegraphics[width=1\linewidth]{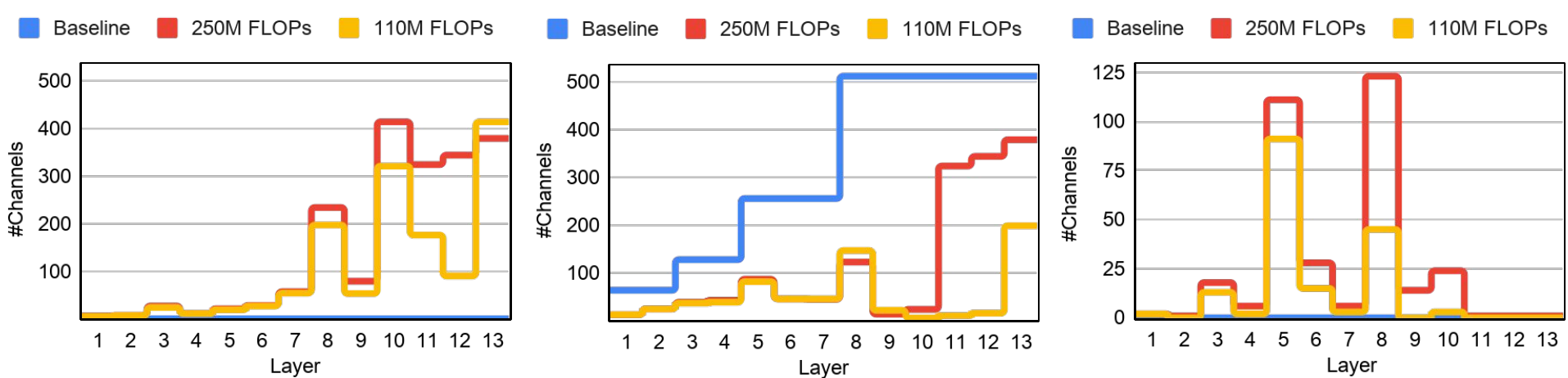} \\ \vspace{-0.5cm}
	 {\small \null \hspace{0.13\linewidth} (a) 1$\times$1 Conv. \hspace{0.225\linewidth} (b) 3$\times$3 Conv. \hspace{0.225\linewidth} (c) 5$\times$5 Conv.  \hspace{\fill}  \null }
	\vspace{0.1cm}
	\caption{Per-layer analysis of Figure~\ref{fig_fgnas_vgg16} (c). 
	(a), (b), and (c) present the number of channels produced by convolutions of 1$\times$1, 3$\times$3, and 5$\times$5 kernel sizes, respectively.
	The baseline network has only convolutions of 3$\times$3 kernel size.
	}  
	\label{fig_fgnas_vgg16_conv_analysis}
	\vspace{-0.2cm}
\end{figure*}

\subsection{Image Classification}
\paragraph{Results on CIFAR-10}
Table~\ref{table:cifar_search} illustrates the performance comparison with the state-of-the-art architectures.
FGNAS~(Large) outperforms the backbone network EfficientNet-B0~\cite{tan2019efficientnet} with smaller number of parameters, and FGNAS~(Small) has 2.1$\times$ smaller parameters than ProxylessNAS-G~\cite{cai2018proxylessnas} with the comparable accuracy.
The search cost of the proposed algorithm is small, but requires more time to find smaller networks.

	\begin{table}[t]
		\centering
		\caption{Channel pruning performance comparison on CIFAR-10.}
		\vspace{0.2cm}\hspace{-0.3cm}
		\scalebox{0.83}{
			\setlength\tabcolsep{3pt}
			\begin{tabular}{lllrr}
				\toprule[1.5pt]
				Model & Method& Type & Top-1 Acc. & FLOPs \\[0.5ex] 
				\midrule[1.0pt]
				\multirow{4}{*}{VGG-16} & Baseline & manual & 93.7 \% & 627 M\\ \cline{2-5}
				& Huang et al.~\cite{Huang2018learning} & policy-gradient & 90.9 \% & 222 M \\
				& Slimming~\cite{liu2017slimming} & rule-based & \textbf{93.6 \%} & 211 M \\
				 & \textbf{FGNAS (P)}& gradient-based & \textbf{93.6 \%} &  \textbf{149 M} \\
				
				\midrule[1.0pt]
				\multirow{4}{*}{VGG-19} & Baseline & manual & 94.0 \% & 797 M \\  \cline{2-5}
				 &  Slimming~\cite{liu2017slimming}   & rule-based & 93.8 \% & 391 M \\
				 & DCP~\cite{zhuang2018dcp}    & gradient-based & 94.2 \% & 398 M \\
				 & \textbf{FGNAS (P)}& gradient-based& \textbf{94.3 \%} & \textbf{348 M} \\
				
				\midrule[1.0pt]				
				\multirow{3}{*}{ResNet-18}&  Baseline & manual & 91.5 \%& 26.0 G \\  \cline{2-5}
				& Huang et al.~\cite{Huang2018learning} & policy-gradient & 90.7 \%&6.2 G\\
				& \textbf{FGNAS (P)}& gradient-based & \textbf{92.5 \%} &  \textbf{1.3 G}\\
				
				\midrule[1.0pt]
				\multirow{3}{*}{ResNet-20}&  Baseline& manual  & 92.2 \% & 81 M \\  \cline{2-5}
				& Soft Filter~\cite{he2018soft} & rule-based  & 91.2 \%&57 M\\
				& \textbf{FGNAS (P)}& gradient-based& \textbf{91.7 \%}& \textbf{34 M}\\
				
				\midrule[1.0pt]
				\multirow{3}{*}{DenseNet-40}&  Baseline& manual  & 94.3 \%& 566 M \\  \cline{2-5}
				& Slimming~\cite{liu2017slimming}  & rule-based& 93.5 \%&188 M\\
				& \textbf{FGNAS (P)} & gradient-based& \textbf{93.6 \%} & \textbf{149 M}\\
				\bottomrule[1.5pt]
			\end{tabular}
		}
		\label{table:pruning}
		\vspace{-0.2cm}
	\end{table}

\begin{table}[t]
	\centering
	\caption{Ablation study of per-channel gating function with VGG-16 on CIFAR-10. 
	Multiple-operation indicates that more than two operations can produce a channel.
	}
	\label{table:ablation_gate}
	\vspace{0.2cm}\hspace{-0.3cm}
	\scalebox{0.86}{
	\setlength\tabcolsep{3pt}
	\begin{tabular}{cccrr } 
        	\toprule[1.5pt]
          Type & Channel Pruning & Multiple-operation & Top-1 Acc. & FLOPs \\[0.5ex] 
	\midrule[1.0pt]
         (1) &  & \checkmark &  91.0 \% & 278 M \\ 
         (2) &  \checkmark & &  91.6 \% & 131 M \\ 
         Ours &   \checkmark & \checkmark &  \textbf{92.5 \%} & \textbf{61 M} \\ 
	\bottomrule[1.5pt]
	\end{tabular}} 
	\\ \vspace{-0.2cm}
\end{table}


\begin{table*}[t]
	\centering
	\caption{The image super-resolution benchmark  for NAS approaches in scaling factor 2. FLOPs is measured to produce a HD image.}
	\vspace{0.2cm}\hspace{-0.25cm}
	\scalebox{0.83}{
	\begin{tabular}{l l c c c c r r} 
		\toprule[1.5pt]
        \multirow{2}{*}{Model}     & \multirow{2}{*}{Type}  & Set5 & Set14 & B100 & Urban100  & \multirow{2}{*}{Parameters} & \multirow{2}{*}{FLOPs} \\  
             &   & (PSNR/SSIM) & (PSNR/SSIM) & (PSNR/SSIM) & (PSNR/SSIM)  &  &  \\ [0.5ex] 
		\midrule[1.0pt] 
		SRCNN~\cite{Dong2014ImageSU} & manual & 36.66 dB / 0.9542 & 32.42 dB / 0.9063 & 31.36 dB / 0.8879 & 29.50 dB / 0.8946 & 57 K & 105.4 G\\ 
		VDSR~\cite{kim2016accurate} & manual	& 37.53 dB / 0.9587 & 33.03 dB / 0.9124 & 31.90 dB / 0.8960 & 30.76 dB / 0.9140 & 665 K & 1,225.2 G\\
		CARN-M~\cite{ahn2018fast} & manual & 37.53 dB / 0.9583 & 33.26 dB / 0.9141 & 31.92 dB / 0.8960 & 31.23 dB / 0.9144 &412 K & 182.4 G\\ 
		CARN~\cite{ahn2018fast} & manual & 37.76 dB / 0.9590 & 33.52 dB / 0.9166 & 32.09 dB / 0.8978 & 31.92 dB / 0.9256 & 1,592 K & 445.6 G\\
		MemNet~\cite{Tai2017MemNet} & manual & 37.78 dB / 0.9597 & 33.28 dB / 0.9142 & 32.08 dB / 0.8978 & 31.51 dB / 0.9312 & 677 K & 5,324.8 G \\
		EDSR~\cite{EDSR} & manual & 38.11 dB / 0.9601 & 33.92 dB / 0.9198 & 32.32 dB / 0.9013 & 32.93 dB / 0.9351 & 40,712 K & 18,769.5 G \\ 
		RDN~\cite{zhang2018residual} & manual & 38.24 dB / 0.9614 & 34.01 dB / 0.9212 & 32.34 dB / 0.9017 & 32.89 dB / 0.9353 & 22,114 K & 10,192.4 G \\
		\midrule[1.0pt]
		FALSR-B~\cite{Chu2019FastAA} & evolution & 37.61 dB / 0.9585 & 33.29 dB / 0.9143 & 31.97 dB / 0.8967 & 31.28 dB / 0.9191 & 326 K & 149.4 G \\
		ESRN-V~\cite{Song2019EfficientRD} & evolution & 37.85 dB / \textbf{0.9600} & 33.42 dB / \textbf{0.9161} & 32.10 dB / \textbf{0.8987} & 31.79 dB / 0.9248 & 324 K & 146.8 G \\
		\hline 
		\textbf{EDSR-FGNAS} & gradient-based & \textbf{37.86 dB} / 0.9593 & \textbf{33.44 dB} / 0.9157 & \textbf{32.11 dB / 0.8987} & \textbf{31.85 dB / 0.9254} & \textbf{212 K} & \textbf{97.6 G}\\
		\bottomrule[1.5pt]
	\end{tabular}}
	\vspace{-0.2cm}
	\label{table:sr_results}
\end{table*}

\vspace{-0.3cm}
\paragraph{Results on ImageNet}
Table~\ref{table:mobilenetv2} presents the performance comparison with MobileNetV2 Multiplier~\cite{sandler2018mobilenetv2} and NetAdapt~\cite{yang2018netadapt}, which successfully prunes channels of efficiently designed networks~\cite{sandler2018mobilenetv2, howard2019mobilenetv3}.
For the fair comparison, we evaluate the proposed algorithm as a channel pruning method, referred as  FGNAS~(P), of which search space is only the number of channels in all layers.
FGNAS~(P) is faster in the both of FLOPs and latency than other channel pruning methods and FGNAS achieves 1.6\% higher Top-1 accuracy than Multiplier.
The model latency reaches the target latency within 40 epochs at the search stage, which indicate the search cost of the proposed algorithm.

\vspace{-0.4cm}
\paragraph{Ablation study of search space}
Our search method easily enlarges search space by adding operations to the layers of backbone networks for more efficient architectures.
Table~\ref{table:space_analysis} shows that the proposed algorithm finds faster networks in large search space with the same Top-1 Accuracy. 
Figure~\ref{figure:pruning_vs_search} draws FLOPs/Accuracy graphs of our search methods.
FGNAS consistently outperforms FANAS~(P) while reducing the network run-time, and finds the 5.7$\times$ smaller FLOPs architecture than original VGG-16 on CIFAR-10.

\vspace{-0.3cm}
\paragraph{Searched architecture analysis}
To analyze the performance improvement from flexible architectures, we visualize two FGNAS architectures, which have 250M and 110M FLOPs from VGG-16 on CIFAR-10.
The search space is 1, 3, 5, 7, 9, and 11 kernel sizes in convolutions and ReLU, PReLU, and tanh in activation functions, and the number of channels in all layers.
The searched networks by FGNAS and original VGG-16 have less than 0.3\% accuracy differences.
Figure~\ref{fig_fgnas_vgg16}~(a) shows that 3, 5, 8, and 10-th layers, located at after pooling operation, remains more channels than next layers and 110M FLOPs network prunes most of channels at 10$\sim$12-th layers of 250M FLOPs network.
As illustrated in Figure~\ref{fig_fgnas_vgg16}~(b), 110M FLOPs network has much higher numbers of operation types within a layer which lead complex layer configurations.
Note that 5-th layer has 31 different operation types.
Figure~\ref{fig_fgnas_vgg16}~(c) shows that 1$\times$1 convolutions appear more frequently for the network efficiency.
Figure~\ref{fig_fgnas_vgg16_conv_analysis}~(a) shows convolutions of 1$\times$1 kernel size produce more channels at 8$\sim$13-th layers, where the feature map resolutions are 4$\times$4 and 2$\times$2 pixels.
On the other hand, 1$\sim$8-th layers prefer 3$\times$3 convolutions than 1$\times$1 and prune most channels at 10-th layer, as illustrated in Figure~\ref{fig_fgnas_vgg16_conv_analysis}~(b).
The channels from convolutions of 5$\times$5 kernel sizes mainly remain at 3, 5, and 8-th layers, located at after pooling operation.

\vspace{-0.3cm}
\paragraph{Channel pruning results on CIFAR-10}
We evaluate the channel pruning performance of our algorithm FGNAS~(P) based on diverse backbone networks of VGGNet~\cite{simonyan2014vgg}, ResNet~\cite{he2015resnet}, and DenseNet~\cite{huang2016dense}. 
Since original standard CNN networks are designed for ImageNet, we adopt the modified networks for CIFAR-10~\cite{liu2017slimming, Huang2018learning}. Table~\ref{table:pruning} shows that the proposed algorithm outperforms the existing pruning methods~\cite{Huang2018learning,liu2017slimming, zhuang2018dcp,he2018soft} even with less FLOPs.
Huang et al.~\cite{Huang2018learning} removes channels layer-by-layer with RL-based policy gradient estimation, of which search cost is 30 GPU days using Nvidia K40.
Since FGNAS~(P) searches over all layers simultaneously using differentiable gating functions, the search cost is 1 GPU hour using GeForce 1080 Ti on CIFAR-10.
We reproduced the DenseNet-40 result of Slimming~\cite{liu2017slimming} for fair comparison.

\vspace{-0.3cm}
\paragraph{Ablation study of gating function}
We evaluate the proposed search algorithm with the modifications of gating function, which exclude its advantages one by one.
Table~\ref{table:ablation_gate} shows that each advantage significantly improves the performance of searched architectures.
Note that Type (2) gating function in Table~\ref{table:ablation_gate} search for an operation per channel, while the gating functions in ProxylessNAS~\cite{cai2018proxylessnas} and FBNet~\cite{Wu2019fbnet} choose one operation per layer.

\subsection{Image Super-Resolution}

To verify the more practical effectiveness of our approach, we evaluate our search method on image super-resolution (SR) tasks.
The primary metric of this task is FLOPs of networks because the FLOPs are easy to calculate regardless of input image resolutions, which are arbitrary in SR problems. 

\vspace{-0.3cm}
\paragraph{Results}
Table~\ref{table:sr_results} shows FLOPs of networks producing an HD image (1280$\times$720 resolution) by scaling factor 2.
Since SR networks require substantially large amount of FLOPs comparing to conventional image classification networks, our search algorithm aims to find faster networks.
FGNAS achieves 1.5$\times$ reduced FLOPs and the number of parameters than the state-of-the-art  NAS approaches~\cite{Chu2019FastAA,Song2019EfficientRD} as illustrated in Table~\ref{table:sr_results}.
Note that FGNAS is even faster than SRCNN~\cite{Dong2014ImageSU}, which consists of 3 convolution layers.
The searched residual blocks have large number of channels and operations for activation.
The number of channels for skip connections gradually increases in the depth of networks.
The search cost is 0.5 GPU day with GeForce 2080 Ti.

\section{Conclusion}
\label{sec:conclusion}
We presented a novel architecture search technique, referred to as FGNAS, which provides a unified framework of structure and operation search via channel pruning.
The proposed approach can be optimized by a gradient-based method, and we formulate a differentiable regularizer of neural networks with respect to resources, which facilitates efficient and stable optimization with the diverse tasks-specific and resource-aware loss functions.



{\small
\bibliographystyle{ieee_fullname}
\bibliography{main}
}

\end{document}